\documentclass[conference]{IEEEtran}
\IEEEoverridecommandlockouts
\usepackage{cite}
\usepackage[table,xcdraw]{xcolor}
\usepackage[colorinlistoftodos,prependcaption,textsize=tiny]{todonotes}
\usepackage{amsmath,amssymb,amsfonts}
\usepackage{algorithmic}
\usepackage{graphicx}
\usepackage{textcomp}
\usepackage{xcolor}
\def\BibTeX{{\rm B\kern-.05em{\sc i\kern-.025em b}\kern-.08em
    T\kern-.1667em\lower.7ex\hbox{E}\kern-.125emX}}

\begin{document}

\title{”Do it my way!”: Impact of Customizations on Trust perceptions in
Human-Robot Collaboration\\
}

\author{\IEEEauthorblockN{Parv Kapoor}
\IEEEauthorblockA{\textit{Institute for Software Research} \\
\textit{Carnegie Mellon University}\\
Pittsburgh, U.S.A. \\
parvk@andrew.cmu.edu}
\and
\IEEEauthorblockN{Simon Chu}
\IEEEauthorblockA{\textit{Institute for Software Research} \\
\textit{Carnegie Mellon University}\\
Pittsburgh, U.S.A. \\
cchu2@andrew.cmu.edu}
\and
\IEEEauthorblockN{Angela Chen}
\IEEEauthorblockA{\textit{Robotics Institute} \\
\textit{Carnegie Mellon University}\\
Pittsburgh, U.S.A. \\
angelac2@andrew.cmu.edu}
}

\maketitle
\thispagestyle{plain}
\pagestyle{plain}

\begin{abstract}
Trust has been shown to be a key factor in effective human-robot collaboration. In the context of assistive robotics, the effect of trust factors on human experience is further pronounced. Personalization of assistive robots is an orthogonal factor positively correlated with robot adoption and user perceptions. In this work, we investigate the relationship between these factors through a within-subjects study (N=17). We provide different levels of customization possibilities over baseline autonomous robot behavior and investigate its impact on trust. Our findings indicate that increased levels of customization was associated with higher trust and comfort perceptions. The assistive robot design process can benefit significantly from our insights for designing trustworthy and customized robots.
\end{abstract}


\section{Introduction}
\label{intro}


\par As robots become more prevalent in society and play an increasingly meaningful
role in our daily lives, trust has become a key factor in effective collaboration
and long-term relationships between humans and robots. It has been shown
that trust increases users’ intention to use and it is negatively correlated with
user anxiety \cite{schule2022patients}. Trust is also a reliable
predictor of task outcome in human-robot collaboration \cite{herse2021using}. Furthermore, the
trust level can provide further insights into the design of future robots for safe and
effective collaboration. \cite{tomzcak2019let}
\par In the assistive/personal robots domain, one of the initial interactions people
have with the robot is customizing its default behavior per their personal
preferences. People tend to curate more personalized experiences through the
customization of the robot, including but not limited to setting safety operating
boundaries for robot manipulators, operation routines, and patterns. Recently, there has been a surge in personalized robots’ demand and they play increasingly pivotal roles in people’s daily lives. With increased human-robot
coexistence, robot designers and engineers realize the need to provide a more
diverse set of customization options for tighter user integration. Even
though the relationship between customization and trust is well-studied in the
context of digital devices and recommendation agents \cite{komiak2006effects}, currently, there’s no
related study in the area of human-robot interaction. We propose the following questions to investigate this relationship. 
\par \textbf{RQ1: Does user customization of robot behavior affect perceived trust in robots' ability for assistive tasks?} 

Additionally, in the field of psychology, A third-person perspective has been shown to regulate one's emotions \cite{wallace2016impact}. While we did not find any related work that links perspective with robotics, various attributes of first-person and third-person perspectives are studied extensively in the area of Virtual Reality (VR). In \cite{debarba2017}, it is stated that the third-person perspective (3PP) is compatible with body ownership when sensorimotor contingencies are present, while the first-person perspective (1PP) can foster the user to develop a strong sense of embodiment. In \cite{gorisse2017}, third-person perspective (3PP) has been shown be to associated with strong spatial awareness and environmental perception capacity, therefore, making the user less vulnerable. Hence, we are also interested in observing the impact of "perspective" on trust. We propose an additional question to explore this relationship.
\par \textbf{RQ2: Does user view perspective affect perceived trust in robots' ability for assistive tasks?}

\par In this paper, we explore user preferences for different customization levels of robotic behavior. We present two studies where 17 participants with various levels of robot use experience interact with a simulated assistive feeding task. Participants were given three levels of control over robot behavior and two perspectives through the simulator. We observe that users perceived the robot as more trustworthy and comfortable in both perspectives with increased levels of control. Additionally, we observe that perspectives did not have a statistically significant impact on trust but had an impact on comfort perceptions.
\section{Related Work}%
\label{related}
\subsection{Trust in Human-Collaboration Research}

Trust in the context of human-robot collaboration has been an important subject of interest\cite{xu2012trust} \cite{sadrfaridpour2016modeling}\cite{krausman2022trust} 
\cite{hancock2011meta}, where some user studies focused on study designs derived from economic and psychology studies \cite{hsieh2020human}\cite{abubshait2021examining}, and some focused on directly measuring users' trust based on an existing robotic design \cite{schule2022patients}\cite{martelaro2016tell}. In most cases, post-study questionnaires are usually used to assess the user's trust level \cite{krausman2022trust}\cite{freedy2007measurement}. However, other measurements of trust have also been explored. For example, Krausman et al. (2022) \cite{krausman2022trust} underlined the different types
of trust and its indicative measures such as subjective, communication,
behavioral, and physiological in a team. In this work, we explored several definitions of trust in psychology \cite{schilke2021} to support our hypotheses. For our work, we will leverage existing work to design our trust elicitation questionnaire and interviews, and use measures such as intervention to indirectly measure trust.

\subsection{Personalization}
Personalization is critical in assistive tasks of daily functions for users with impairments \cite{alqasemi2005wheelchair}. It has been specifically investigated in the context of assistive feeding in \cite{bhattacharjee2020more}\cite{bhattacharjee2019towards}. In  Bhattacharjee et al (2020) \cite{bhattacharjee2020more} studied user preferences for different levels of autonomy in feeding. Gordon et al (2020) \cite{gordon2020adaptive} created a simulated environment that adapts to user food preferences. In another line of work, Schultz et al (2022) \cite{schultz2022proof} presented a prototype robot-assisted self-feeding system that takes the eye and head movements for food preferences.

However, existing literature has not explored the relationship between trust and the level of customization in assistive robot applications like assistive feeding and drinking. To simplify the problem, We identified the key factor that directly affect the users' experience interacting with the robot: the level of control. Additionally, we also investigate the effect of viewing perspectives on trust perceptions. Neither of these have been studied in the context of trust in human-robot interactions. There is a large body of work in the area of virtual reality (VR), which studies the relationship between vulnerability, sense of embodiment, ownership, agency, and viewing perspectives \cite{debarba2017} \cite{gorisse2017}. We leveraged this work in viewing perspectives and the definition of trust in psychology to propose our hypothesis in the following sections, and explored the relationship between different levels of control, viewing perspectives, and trust.

\section{System Overview}%
\label{methods}
\subsection{Simulator}
We use an assistive feeding task for our study. We modify a preexisting simulation environment (assistive gym \cite{erickson2020assistivegym}) which is a state of the art physics based simulator for physical robotic assistance. In order to make the scenarios hyper realistic we use textures from iGibson \cite{li2022igibson} simulator. We load the indoor household scenario from iGibson to mimic the traditional setup of a participant's household. We initialize the Jaco 2 robotic arm holding a spoon inside the environment. We assume that the food acquisition stage has already been performed. The task is succesful if the robotic arm is able to maneuver the spoon to the participant's mouth without dropping any simulated food particles. 
\subsection{Modes of Control} 

We train a baseline policy for autonomous control using proximal policy optimization (PPO) \cite{schulman2017proximal} algorithm which is a state of the art reinforcement learning algorithm for continuous action spaces. The environment provides high reward if all food particles are successfully fed to the human agent in the shortest time possible. There is a negative reward associated with dropping food particles from the spoon. 

We discretize the level of control over baseline policy into three categories with increasing levels of complexity. The inputs from the participants are taken using the keyboard.
\begin{enumerate}
    \item \textbf{No intervention control}: the participant has no control over baseline policy and the whole setup is fully autonomous 
    \item \textbf{Stop control}: the participant can stop the robot agent at any time by pressing 'S' and the arm will retract to its initial position. The participant can then press 'R' to resume and the agent will continue trying to feed the human agent. 
    \item \textbf{Granular control}: The participant can perform stop action but in this case also adjust the robot arm's position while its stopped by pressing one of the six designated keys to move it up, down, left, right, front and back. Once content, the participant can then resume the autonomous policy from the new adjusted position. 
\end{enumerate}
\subsection{Perspective} 
We investigated two main perspectives in our study to identify their impact on trust. 
\begin{enumerate}
    \item \textbf{Third Person Perspective (TPP)}: In this situation, we initialise the simulation view as a third person view of the robot agent feeding a human agent sat in the chair as illustrated in figure 4. We ask participants to imagine themselves as a caretaker for this human agent, monitoring robot behavior. \\
    \item \textbf{First Person Perspective (FPP)}: In this situation, we initialise the simulation view as a first person view of the robot agent feeding the human agent as illustrated in figure 5. We asked the participants to imagine themselves as the human agent being fed by the robot. 
\end{enumerate}

\section{User Study}%
\label{methods}
\begin{figure*}
\centering
\includegraphics[scale=0.3]{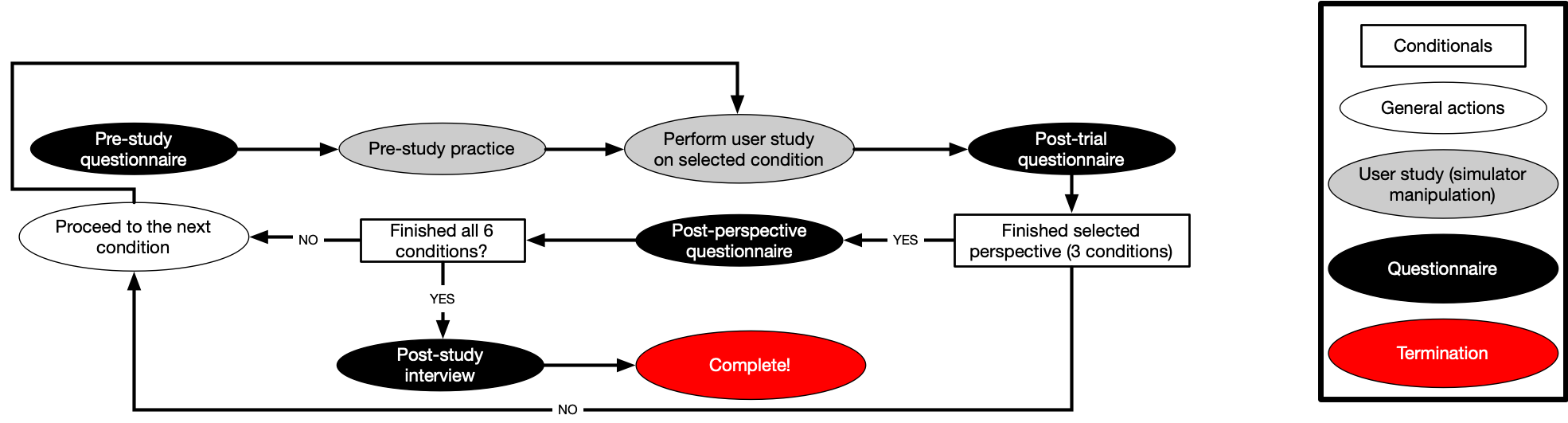}
\caption{Flow chart for the user study}
\label{fig:flowchart}
\end{figure*}

\begin{table}[h]
\caption{Participant demographics information}
\label{table_I}
\begin{center}
\begin{tabular}{|c||c||c|}
\hline
Participant & Age & Expertise\\
\hline
P1 & 27 & Software Engineering \\
P2 & 29 & Software Engineering \\
P3 & 23 & Software Engineering \\
P4 & 22 & Software Engineering \\
P5 & 24 & Health Care \\
P6 & 44 & Software Engineering \\
P7 & 27 & Software Engineering \\
P8 & 29 & Human-Robot Interaction \\
P9 & 25 & Software Engineering \\
P10 & 23 & Software Engineering \\
P11 & 23 & Societal Computing \\
P12 & 23 & Software Engineering \\
P13 & 24 & Human-Computer Interaction \\
P14 & 26 & Human-Computer Interaction \\
P15 & 23 & Human-Computer Interaction \\
P16 & 23 & Software Engineering \\
P17 & 23 & Software Engineering \\
\hline
\end{tabular}
\end{center}
\end{table}



\subsection{Participants}
Seventeen participants (N=17) were recruited from Carnegie Mellon University Robotics Institute (RI), Institute for Software Research (ISR), and University of Pittsburgh Medical Center (UPMC). The study is a within-subject study. The participants have expertise spanning from software engineering, human-computer interaction, human-robot interaction and health care. Over half of the participants (9 out of 17) have previous experience interacting with a robot. 7 of the participants have direct or indirect experiences interacting with assistive devices. 



\begin{figure}[h!]
\includegraphics[width=0.9\columnwidth]{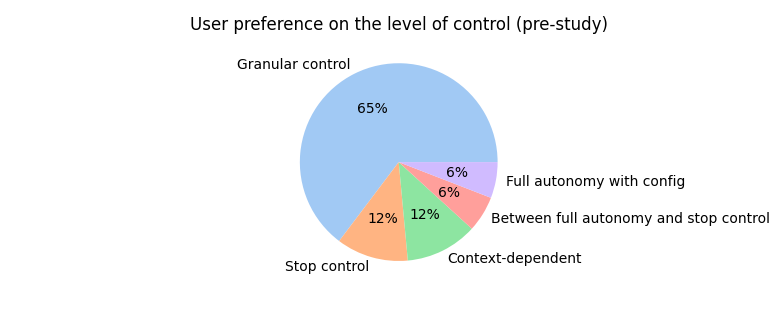}
\caption{Preferred level of autonomy in pre-study questionnaire}
\label{fig:FPV}
\end{figure}
\subsection{Design and Procedure}
\begin{table}[h]
\caption{Independent Variables of the User Study}
\label{table_I}
\begin{center}
\begin{tabular}{|c||c|}
\hline
Control & Perspectives\\
\hline
No Intervention Permitted & First-person View \\
Stop Action & 3rd-person View \\
Stop Action and Granular Control & \\
\hline
\end{tabular}
\end{center}
\end{table}
\subsubsection{Design}
The experiment was conducted in a controlled environment with participants facing the wall and the researchers instead of the windows to mitigate biases caused by environmental stimulation. The study was conducted indoors in a conference room at TCS hall, Carnegie Mellon University.
As presented in Table \ref{table_I}, we identified 2 independent variables for the case study. (1) the level of control over the robot's behavior and (2) the view perspectives. We have two dependent variables (1) the self-reported trust level for each trial condition and (2) the self-reported comfort level, which is hypothesized as a latent variable for the perceived trust.\\

\noindent{\textbf{Quantitative measurement}}: The quantitative measurements we are collecting are self-reported trust and comfort levels on a Likert scale. We also asked the participants to rank the level of different control levels after experiments with each perspective, and overall after all the trials, and rate their most preferred autonomy level and perspectives. In addition, we also ask the participants question "would you consider using our system again in the future?" in the post-study. The yes/no response of the latter question serves as a latent variable for the quality of the implementation of the system.

\noindent{\textbf{Qualitative measurement}}: The qualitative measures here we are collecting are divided into 3 parts (1) pre-study (2) post-trial and (3) post-study. In the pre-study, we ask participants to elaborate on their previous experience with robotics applications and assistive devices. In the post-trial survey, we try to elicit comments on their experiences with the particular condition. In the post-study questionnaire, we elicited the reason why the participants will use the system again in the future, and their feedback on improving the experience design and setup.


Our first and second hypotheses are:\\


\noindent{\textbf{H1: Trust is influenced by the level of participant control.}}


\noindent{\textbf{H2: Comfort is influenced by the level of participant control.}}

Since the related work does show that the first-person view makes people more vulnerable, and the nature of assistive tasks fall is well suited for these situations. In lieu of this, We state our third and fourth hypotheses:\\

\noindent{\textbf{H3: The change of perspective has an impact on trust.}}\\

\noindent{\textbf{H4: The change of perspective has an impact on comfort.}}

Intuitively, this is due to the case that the third-person perspective enables the participants to be less vulnerable and more detached from the assistive context. \\

We have explored latent variables of trust, and previous work has indicated that comfort level can be a latent variable of trust, therefore, we have a fifth hypothesis: \\

\noindent{\textbf{H5: Higher comfort levels are correlated with higher trust levels}}



\begin{figure}[h!]
\centering
\includegraphics[width=0.9\columnwidth]{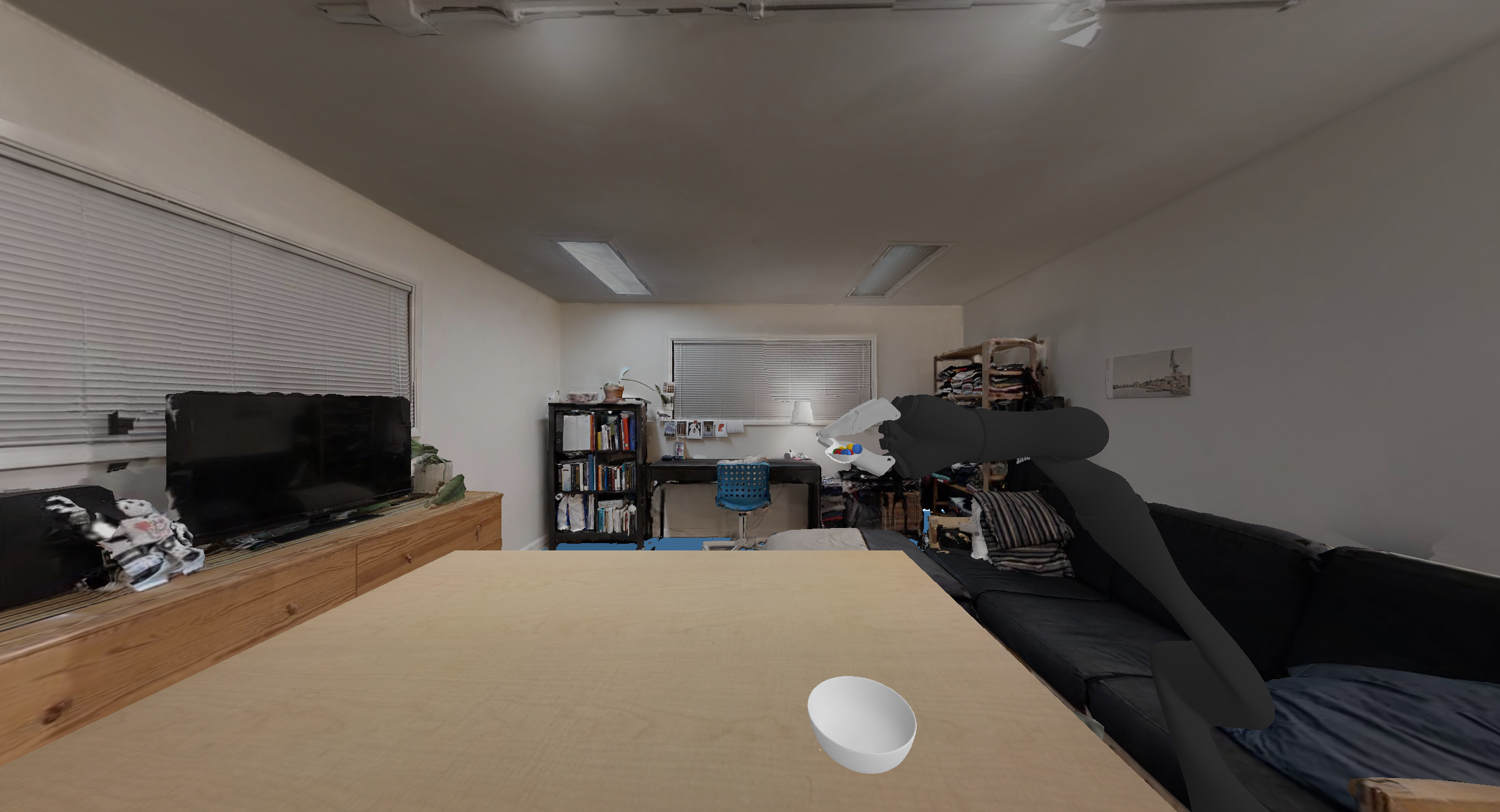}
\caption{First-person view of the feeding task in the simulator}
\label{fig:FPV}
\end{figure}

\begin{figure}[h!]
\centering
\includegraphics[width=0.9\columnwidth]{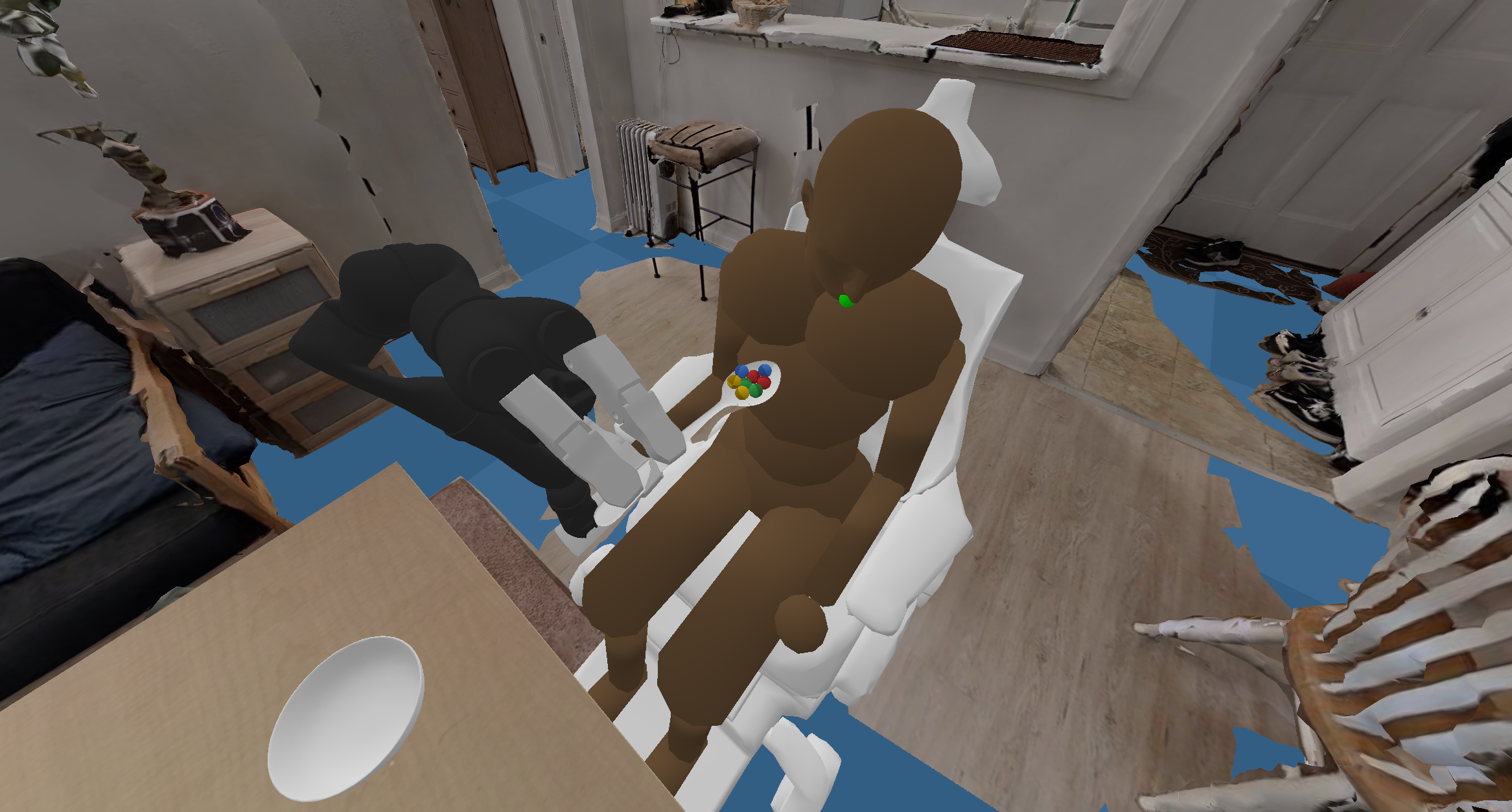}
\caption{Third-person view of the feeding task in the simulator}
\label{fig:TPV}
\end{figure}
We investigated 3 conditions for the level of control, 2 conditions for perspectives for the assistive feeding task. This lead to total of 6 scenarios. Additionally, we simulate 3 episodes per scenario to collect more data points and account for any bias. The conditions and parameters are summarized in table \ref{table_I}. 





\subsubsection{Procedure} 
The procedure is shown in figure \ref{fig:flowchart}. After welcoming the participants, each experiment session started with signing the consent form, demographic questionnaire, and a short introduction to the study. The questionnaire given prior to the study has three goals (1) to elicit demographic information (2) to understand prior experience interacting with robots and (3) to record study participants' preferred level of control. For the demographics questionnaire, we will ask questions like gender, age, education level, and area of study. 

After the pre-study questionnaire, a pretrial was conducted  to get the participants accustomed to the user interface and the simulator. We let the participants teleoperate with the robotic arm for feeding and drinking tasks using the keyboard. 

The post-study questionnaire was given (1) after each trial and (2) after the completion of the entire user study. After each trial, the participants will be given a Likert-scale questionnaire in which they can rate the trust and comfort level that they feel towards the robot. Upon the completion of the study, the participants will be asked to pick the conditions with the highest trust and comfort levels.   

A 15-minute semi-structured interview was conducted to conclude the study, mainly to elicit user feedback for the user study.

\section{Results and Evaluation}%
\label{eval}

\subsection{Descriptive statistics}

All analyses were performed using statistical packages like \cite{seabold2010statsmodels} in Python. The data set consists of 90 data points (15 participants x 6 trials). A descriptive statistics for trust score and comfort score are as follows. 

\begin{table}[h]
\begin{center}
\caption{Descriptive statistics of the trust scores and comfort scores}
\begin{tabular}{|c|c|c|c|}
\hline
Scores & N & Mean & SD\\
\hline
Trust & 90 & 3.044444 & 1.271212 \\
  Comfort   & 90 & 2.822222 & 1.285859 \\
\hline
\end{tabular}
\end{center}
\end{table}

The descriptive statistics of the trust score group by the control level are shown in TABLE III . A similar set of descriptive statistics of the comfort score are shown in TABLE 4. 

\begin{table}[h]
\begin{center}
\caption{Descriptive statistics of the trust scores grouped by various levels of autonomy level}
\begin{tabular}{|c|c|c|c|}
\hline
Control & N & Mean & SD\\
\hline
no & 30 & 2.3667 & 1.2452 \\
stop     & 30 & 3.3000 & 1.1492 \\
granular & 30 & 3.4667 & 1.1666 \\
\hline
\end{tabular}
\end{center}
\end{table}

\begin{table}[h]
\begin{center}
\caption{Descriptive statistics of the comfort scores grouped by various levels of autonomy level}
\begin{tabular}{|c|c|c|c|}
\hline
Control & N & Mean & SD\\
\hline
no & 30 & 1.8333 & 0.9855 \\
stop     & 30 & 3.2000 & 1.1567 \\
granular & 30 & 3.4333 & 1.1043 \\
\hline
\end{tabular}
\end{center}
\end{table}

The descriptive statistics (mean and standard deviation) of the trust score grouped by perspective are shown in TABLE 5. A similar set of descriptive statistics of the comfort score are shown in TABLE 6. 

\begin{table}[h]
\begin{center}
\caption{Descriptive statistics of the trust scores grouped by various levels of perspective}
\begin{tabular}{|c|c|c|c|}
\hline
Perspective & N & Mean & SD\\
\hline
1st & 45 & 2.8667 & 1.2898 \\
3rd & 45 & 3.2222 & 1.2411 \\
\hline
\end{tabular}
\end{center}
\end{table}

\begin{table}[h]
\begin{center}
\caption{Descriptive statistics of the comfort scores grouped by various levels of perspective}
\begin{tabular}{|c|c|c|c|}
\hline
Perspective & N & Mean & SD\\
\hline
1st & 45 & 2.6000 & 1.3212 \\
3rd & 45 & 3.0444 & 1.2239 \\
\hline
\end{tabular}
\end{center}
\end{table}

\subsection{Statistical testings}

To test Hypothesis H1-H4, we performed a repeated measure two-way ANOVA test on the trust score (Likert scale 1-5) and comfort score (Likert scale 1-5). The independent variables are control (three levels: no, stop, and granular) and perspective (two levels: 1st and 3rd). 

The repeated measure ANOVA test for trust score shows control (F statistics = 28, p = 0.0001) has a statistically significant effect on trust under alpha = 0.05, whereas perspective (F statistics = 14 , p = 0.0757) doesn't have enough evidence to suggest it makes a significant effect on trust under alpha = 0.05. The interactive effect (F statistics = 28 , p = 0.2129) between control and perspective also concludes no significant effect under alpha = 0.05. 

Fig. 5 depicts the relationship between two different perspectives fixed at each control level based on the mean trust score. We observe a general upward trend of trust as the level of control increases, while the discrepancy of trust closes as the level of control increases. 

Fig. 6 depicts the relationship between three different control levels fixed at two different perspectives based on the mean trust score. We observe a general upward trend of trust as the perspective changes from the first person to the third person.

The repeated measure ANOVA test for comfort score shows control (F statistics = 28, p = 0.0000) has a statistically significant effect on comfort under alpha = 0.05, and perspective (F statistics = 14 , p = 0.0058) also has a statistically significant effect on comfort under alpha = 0.05. However, the interactive effect (F statistics = 28 , p = 0.1823) between control and perspective doesn't have enough evidence to suggest it makes a significant effect on trust under alpha = 0.05. 

Fig. 7 depicts the relationship between two different perspectives fixed at each control level based on the mean comfort score. We observe an upward trend of comfort at the first person perspective. However, at the third person perspective, we notice an interesting drop of comfort as the control becomes more complicated ($stop \longrightarrow granular$).

Fig. 8 depicts the relationship between three different control levels fixed at two different perspectives based on the mean comfort score. We observe a general upward trend of comfort as the perspective changes from the first person view to the third person view.

To test Hypothesis H5, we ran Pearson's correlation test. Our test results (statistic=0.720, pvalue=1.314e-15) shows the two variables are positively and significantly correlated under the alpha threshold 0.05. 

To summarize, 
our \textbf{hypothesis H1 was correct} as the p-value (p = 0.0001) that we obtained through the two-way repeated ANOVA test between the control level and trust demonstrates participant control has a significant effect on trust. 

Our \textbf{hypothesis H2 was correct}, as the p-value (p = 0.0000) that we obtained through the two-way repeated ANOVA test between the control level and comfort demonstrates participant control has a significant effect on comfort.

Our \textbf{hypothesis H3 was incorrect}, as the p-value (p = 0.0757) that we obtained through the two-way repeated ANOVA test between perspective and trust demonstrates perspective does not have a significant effect on trust.

Our \textbf{hypothesis H4 was correct}, as the p-value (p = 0.0058) that we obtained through the two-way repeated ANOVA test between perspective and comfort demonstrates perspective has a significant effect on comfort.

Our \textbf{hypothesis H5 was correct}, as the p-value (p = 1.314e-15) that we obtained through the correlation test between the comfort level and the trust level demonstrates that there's a statistical significance between the correlation.

\begin{table}[h]
\begin{center}
\caption{Two-way ANOVA table on Trust}
\begin{tabular}{|c|c|c|c|c|}
\hline
& sum of sq & df & F & PR\\
\hline
Control      &       13.6828 & 2.0000 & 28.0000 & 0.0001\\
Perspective    &      3.6797 & 1.0000 & 14.0000 & 0.0757\\ ControlXPerspective & 1.6354 & 2.0000 & 28.0000 & 0.2129 \\

\hline
\end{tabular}
\end{center}
\end{table}

\begin{table}[h]
\begin{center}
\caption{Two-way ANOVA table on Comfort}
\begin{tabular}{|c|c|c|c|c|}
\hline
& sum of sq & df & F & PR\\
\hline
Control       &      38.7885 & 2.0000 & 28.0000 & 0.0000\\
Perspective    &     10.5660 & 1.0000 & 14.0000 & 0.0058\\
ControlXPerspective  & 1.8097 & 2.0000 & 28.0000 & 0.1823\\

\hline
\end{tabular}
\end{center}
\end{table}

\begin{figure}
     \centering
    \includegraphics[width=0.9\columnwidth]{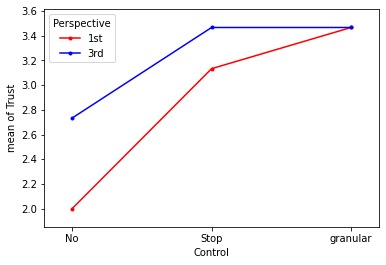}    
      \caption{Mean trust grouped by control} \label{fig:anova}
\end{figure}

\begin{figure}
     \centering
    \includegraphics[width=0.9\columnwidth]{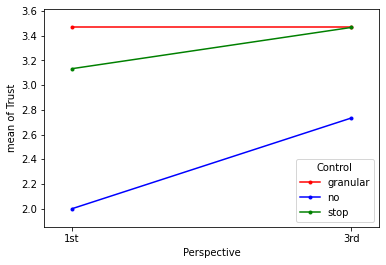}    
      \caption{Mean trust grouped by perspective} \label{fig:anova2}
\end{figure}

\begin{figure}
     \centering
    \includegraphics[width=0.9\columnwidth]{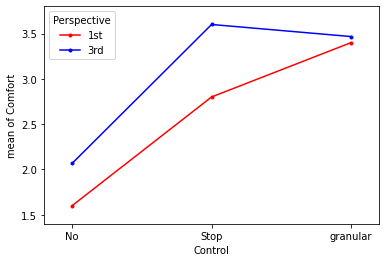}    
      \caption{Mean comfort grouped by control} \label{fig:anova3}
\end{figure}

\begin{figure}
     \centering
    \includegraphics[width=0.9\columnwidth]{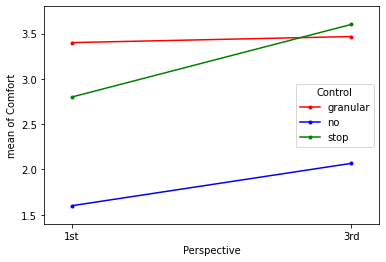}    
      \caption{Mean comfort grouped by perspective} \label{fig:anova4}
\end{figure}





\section{Discussion}%
\label{discuss}

\subsection{Post-study quantitative feedback analysis}
Overall 13 participants (77\%) would use the system again in the future. A majority (65 percent) of our participants chose to have more granular control over a robot in our pre-study. After being exposed to six different control/perspective conditions, 70 percent of our participant chose to have more granular control. Roughly one third of the participants who chose granular control changed their answers. Some of their reasonings include "granular control was not intuitive" and "it is difficult to control the robot through simulations and the keyboard". The qualitative responses are insightful for us to construct our future study, where we could consider an alternative input and interface that is not through a keyboard and simulations. Instead, we could deploy a real robot (e.g., Jaco) and consider eye gaze and verbal cues for inputs. 

Since we exposed our participants to the same six conditions, we performed repeated ANOVA on our data. However, even though our within subject study design allows us to analyze with more sample size (higher statistical power), we acknowledge that our analysis have the following short comings: (1) we treated our response as a continuous variable and made the assumption that it follows a normal distribution (2) our sample size is still too small. Due to the shortcomings, we believe the statistical results are not completely conclusive. Therefore, the qualitative feedback we got from our surveys were critical for us to validate our statistical results. Furthermore, the descriptive statistics (table 3-7) and the visualization of the trust and comfort scores (fig 5-8) provided more context for us to interpret our statistical results. For example, more control was rated with a higher trust score and comfort score on average. Therefore, it makes sense that control has an effect on either trust and comfort.

\subsection{Post-study qualitative feedback analysis}
Finally, based on the post-study interview, we found out that the decline in trust level was caused by mainly 3 reasons. (1) cognitive overload: granular control has 3 degrees of freedom, which results in a sudden increase in the cognitive load. (2) difficulty in seeing the relative position due to control perspective: during the experiment, the control key shortcut is designed based on the 1st person's perspective. Controlling the robot from a 3rd-person perspective requires doing the opposite of the control shortcuts. This increased difficulty may be contributing to the decrease in comfort (3) Concern about control hurting a 3rd person: Due to the fact in the 3rd-person perspective, the are liabilities and responsibilities for the participant to operate correctly instead in full autonomy and stop control, and liability is largely placed on the robot.





%

\begin{figure*}
\centering
\includegraphics[scale=0.1]{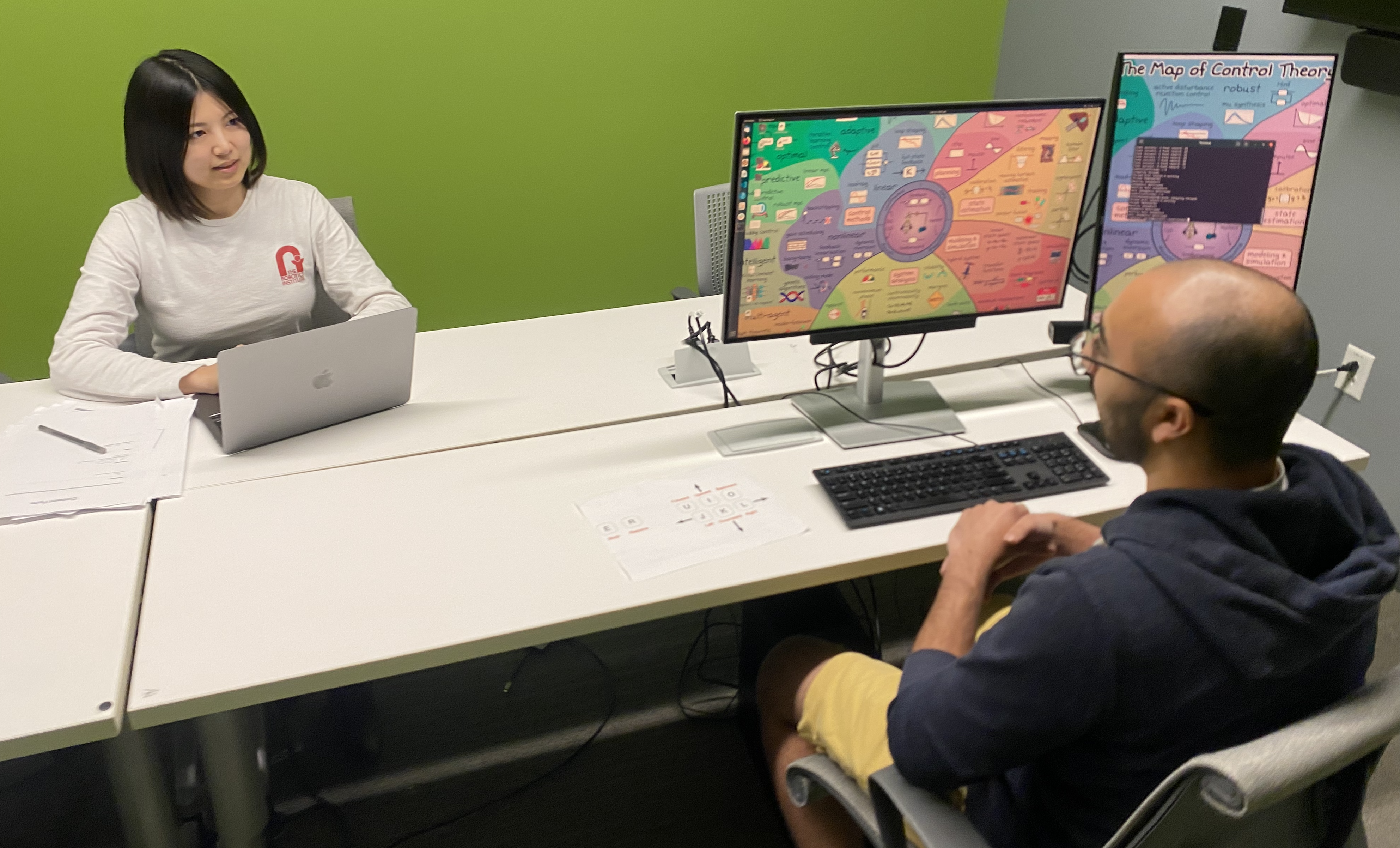}
\caption{The setup of a user study scenario}
\label{fig:flowchart}
\end{figure*}

\section{Future Work}
\label{future-work}
\textbf{Physical robot experiments}: We would like to conduct this user study again with a physical Jaco Robot, which increases the realism and conforms to the actual deployment scenario. We aim to use the lessons learned from running a study through simulators and find more interesting results through study redesign.

\textbf{Anthropomorphism}: During the user study, we have feedback from P6 regarding anthropomorphism, the subject mentioned that the robot looks more like an industrial robot rather than an assistive feeding robot which makes them feel uncomfortable. Adding an anthropomorphism component to the robotic arm may make the robot more approachable.

\textbf{Feedback mechanisms}: During the case study, we have feedback from P14 about the lack of diverse control input and feedback mechanisms, especially as the user of the assistive feeding robot may have mobility issues. In the future, in addition to the keyboard input mechanism, we would like to explore other input/feedback mechanisms, such as using eye gaze/blinking, and head tilting,  to fulfill diverse use cases. In the current setup (especially in the case of no control), the robot has no way of processing feedback mechanisms from the users, and the lack of communication results in perceived parent-child-like power dynamics between the robot and the patient. We hypothesize that the additional feedback mechanism can make the human feel heard and respected, and preserve their dignity.

\section{Conclusion}
\label{conclusion}
To study the relationship between the control level given to the user and the perceived trust, we conducted a user study with 17 participants (N=17) from various disciplines using an assistive feeding task. Through our experiment, we used a mix-methods research methodology with surveys and interviews and discovered based on the statistical analysis that the more control given to the user, the more trust and comfort the user has towards the robot. Perceived comfort has been shown to have a significant correlation with perspective but perceived trust does not have a significant correlation with perspective. Trust is shown to be correlated with comfort level, indicating that comfort level can be an accurate predictor of trust. Finally, the 1st perspective has achieved a consistently lower level of trust than the 3rd person perspective. 

\bibliographystyle{IEEEtran}

\bibliography{references}

\end{document}